\documentclass[conference]{IEEEtran}
\IEEEoverridecommandlockouts
\usepackage{cite}
\usepackage{amsmath,amssymb,amsfonts}
\usepackage{algorithmic}
\usepackage{graphicx}
\usepackage{textcomp}
\usepackage{xcolor}
\def\BibTeX{{\rm B\kern-.05em{\sc i\kern-.025em b}\kern-.08em
    T\kern-.1667em\lower.7ex\hbox{E}\kern-.125emX}}

\begin{document}

\title{Improving Neural Network Learning Through Dual Variable Learning Rates}

\author{\IEEEauthorblockN{Elizabeth Liner}
\IEEEauthorblockA{\textit{Department of Computer Science} \\
\textit{The University of Texas at Austin}\\
Austin, TX, USA \\
e.liner@utexas.edu}
\and
\IEEEauthorblockN{Risto Miikkulainen}
\IEEEauthorblockA{\textit{Department of Computer Science} \\
\textit{The University of Texas at Austin}\\
Austin, TX, USA \\
risto@cs.utexas.edu}
}

\maketitle


\begin{abstract}
This paper introduces and evaluates a novel training method for neural networks: Dual Variable Learning Rates (DVLR). Building on insights from behavioral psychology, the dual learning rates are used to emphasize correct and incorrect responses differently, thereby making the feedback to the network more specific. Further, the learning rates are varied as a function of the network's performance, thereby making it more efficient. DVLR was implemented on three types of networks: feedforward, convolutional, and residual, and two domains: MNIST and CIFAR-10. The results suggest a consistently improved accuracy, demonstrating that DVLR is a promising, psychologically motivated technique for training neural network models.
\end{abstract}

\section{Introduction}
Behavioral psychology focuses on how humans and animals behave and how behavior drives learning and growth. By bringing such insights into machine learning, it may be possible to create methods that train an artificial neural network in a similar manner to how humans and animals are trained, potentially resulting in better performance  of the neural network.

This paper proposes such a method: dual variable learning rates, or DVLR. Dual learning rates provide different emphasis for correct and incorrect responses and thus propagate more specific feedback to the network. The learning rates are updated with a variable rate of change based on the performance of the network so that feedback can be used most efficiently over time. This novel training technique was tested on the MNIST and CIFAR-10 databases with three different types of architectures: feedforward, convolutional, and residual. The results suggest a consistently improved accuracy in both tasks and all three networks, suggesting that it can serve as a general technique for improving neural network learning.

The paper begins by reviewing the behavioral psychology foundation for the DVLR method in the Background section as well as Related Work on variable learning rates. In the Method section, the specific differences between DVLR and backpropagation are discussed. The Baselines and Thresholds section presents the preliminary experiments in configuring DVLR, and the Results section analyzes how the method performed on the MNIST and CIFAR-10 databases. The Discussion section evaluates the significance of results given the computational complexity of deep learning experiments, and the possibility of constraining the method with biological insights.

\section*{Background}

Behavioral Psychology determines how a subject learns by observing the subject's behavior instead of attempting to explain the subject's thought process. Learning is seen as an enduring change in the mechanisms of behavior as distinct stimuli are paired with responses that result from prior experience. Through experiments, behavioral psychologists can identify what the subject is capable of learning, and the best ways to facilitate or inhibit that learning. This focus on behavior is key to the DVLR method. Computer scientists do not fully understand why a neural network produces the responses it does, especially as networks become more complicated. It is thus difficult to determine what needs to change in the neural network to increase accuracy.  Building on Behavioral Psychology, DVLR attempts to use a network's behavior to create more efficient learning and increase the accuracy of the network.

The origins of this approach can be traced to psychologists Edward Thorndike and B.F. Skinner. Thorndike studied animal intelligence with the use of puzzle boxes and determined that every response of an animal is the result of an interaction with the environment \cite{thorndike}. He rejected randomness in animal actions, and determined that they must be able to form associations just as humans do. His Law of Effect states that the satisfaction or dissatisfaction that the animal receives from an action it performs determines directly if the animal will perform that action again. If the result of an action is favorable, the animal is more likely to perform it; if the result of an action is unfavorable, the animal is less likely to perform it. By providing both favorable and unfavorable feedback to an animal subject, it is possible to teach it to perform or not perform certain actions.

Skinner studied how subjects perform with reinforcement over time, and how various schedules affect the subject's performance \cite{skinner}. Through experimentation, Skinner defined four different types of schedules: fixed ratio, variable ratio, fixed interval, and variable interval. In the variable ratio (VR) schedule of reinforcement, a subject is reinforced after a variable number of responses. Skinner concluded that VR schedules led to the subjects accurately performing tasks faster and for a longer continuous period of time than their counterparts on other schedules.

Similarly in neural networks, the goal is to create the most efficient and accurate networks to solve a specific problem. Building on Thorndike's Law of Effect, separate feedback can be provided for correct and incorrect responses. Building on Skinner's variable ratio schedule, the separate learning rates can be updated after a variable number of responses to change the amount of emphasis a correct or incorrect response has on training. The resulting technique, DVLR, implements these ideas as dual learning rates on variable schedules, as will be discussed next.

\section*{Related Work}

There is prior computational work in using different learning rates for the different parameters of a neural network. For example, Kim, Cho and Lee \cite{kim} assigned a distinct learning rate to each reference vector in their vector quantization model and updated the reference vectors with a competitive learning method. The networks performed faster and more accurately when using more than one learning rate for the network. The main difference from DVLR is that their method uses one learning rate for each reference vector, which increases the number of parameters significantly.

On the other hand, Smith \cite{smith} used a non-stationary learning rate that cycles between reasonable boundary values. He was able to achieve a significant increase in accuracy for the CIFAR-10 domain. DVLR takes this idea one step further by introducing insights from behavioral psychology to determine how the learning rates should change as a function of its performance.

\section*{Method}

DVLR is an extension of the standard gradient descent update method in neural networks \cite{backprop}. There are two key changes that will be discussed in detail: dual learning rates and learning rate updates.

\subsection*{Dual Learning Rates}
In DVLR, two learning rates are used: $\eta_\mathrm{C}$ for correct responses, and $\eta_\mathrm{I}$ for incorrect responses. By splitting up the correct and incorrect responses, it is possible to provide different feedback to the network based on whether its responses were favorable or unfavorable. The hypothesis is that the network will receive more specific feedback and in turn, will learn the ideal weight values more efficiently.

To make the dual learning rate implementation practical, batching was used, where batched responses are a mixture of correct and incorrect responses. Theoretically, the correct or incorrect learning rate would be determined for each response, but this approach is computationally expensive and does not provide a major advantage based on preliminary experiments. Instead, if the majority of responses in a batch are correct, $\eta_\mathrm{C}$ is used and if the majority of responses in a batch are incorrect, $\eta_\mathrm{I}$ is used. 

\subsection*{Learning Rate Updates}
In a network using backpropagation, the learning rate determines the amount of emphasis the error has on the network's weight update. For DVLR, the amount of emphasis changes over time as the learning rate is updated. The hypothesis is that in this manner, the network might discover nuances in the data that were not previously apparent and thus, more accurate networks should result. In preliminary experiments, several types of changes in $\eta_\mathrm{C}$ and $\eta_\mathrm{I}$ were evaluated. The learning rates were changed in different amounts, and after different numbers of correct and incorrect responses had been observed. More specifically, a variable threshold was implemented by varying how many correct and incorrect responses needed to be observed before the learning rate was changed. 

\begin{figure}[t]
\centerline{\fbox{\includegraphics[scale=0.75]{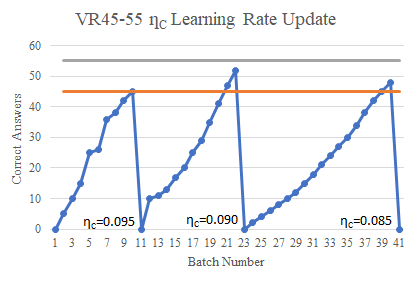}}}
\caption{An example update schedule for the $\eta_\mathrm{C}$ learning rate over three updates ($\eta_\mathrm{I}$ is adjusted in a similar process). The red and grey lines are the boundaries of the threshold, and the blue line is the running count of correct responses from the network. The learning rate changes (i.e.\ decreases by 0.005 in this case) each time the number of correct responses since the last change reaches the threshold. The next threshold is chosen randomly within the range, thus implementing the idea of variable thresholds for correct and incorrect responses.}
\vspace{-2ex}
\end{figure}

The conclusion was that the best performance resulted from variable thresholds and constant rates of change. In DVLR, once the number of correct or incorrect responses reaches the current threshold, the learning rate is updated with the constant rate of change. This method is similar to the variable ratio (VR) schedule in behavioral psychology with one difference: In VR, reinforcement is only given once the subject reaches the threshold, whereas in DVLR, feedback (in the form of gradient) is provided after every example. This difference is due to the inherent nature of neural networks: if gradients were not provided for every example, they would not have any influence on learning.

An example of a learning rate update is shown in Figure~1 for $\eta_\mathrm{C}$; an analogous method is used for $\eta_\mathrm{I}$. A random number within a range (45-55 in this example) is chosen as the threshold. As the network works through examples from the dataset, the number of correct responses is counted. Then, once this number reaches the threshold, the learning rate is updated with the constant rate of change (0.005 in this example), the count is reset to zero, and a new threshold is randomly chosen within the range. This update method continues for the entire span of the experiment. In Figure 1, the learning rate was decreased, but the update direction and magnitude varied in the DVLR experiments, as described in the next section.

\begin{figure}[t]
\centerline{\fbox{\includegraphics[scale=0.75]{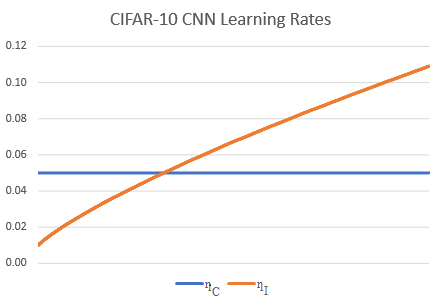}}}
\caption{Example of the learning rates obtained during one trial of the CIFAR-10 CNN model experiments where $\eta^0_\mathrm{C}=0.05$, $\eta^0_\mathrm{I}=0.01$, VT175-225, 0.01\% increase. For the CIFAR-10 CNN, DVLR performed the best when $\eta_ \mathrm{C}$ was static, and when $\eta_ \mathrm{I}$ increased at a nearly constant rate.}
\end{figure}

\section*{Baselines and Thresholds}
The section details the methodology used for determining the best DVLR thresholds. Eventually it may be possible to develop a theoretical approach; however currently the best way to discover them is empirically with a series of experiments. All experiments in this section included three trials to determine the average test accuracy, unless otherwise specified.

Two different baselines were utilized to evaluate the effectiveness of DVLR. The Static-Simple (S-S) Baseline uses the standard, simple learning rate, $\eta_\mathrm{S}$, of normal gradient descent. The Static-Dual (S-D) Baseline has two static learning rates: the rate for correct responses, $\eta_\mathrm{C}$, and the rate for incorrect responses, $\eta_\mathrm{I}$. The ideal learning rate for the S-S baseline was determined empirically as shown in the first half of Table A.III in the Appendix. To discover the static-dual (S-D) baseline, two learning rates were tested around the ideal S-S rate at various distances, as shown in the second half of Table A.III. 

The following notation is used to specify each experiment: The initial learning rate is specified as $\eta^0_ \mathrm{C}=x$ and $\eta^0_\mathrm{I}=y$. The variable threshold is given as VT $\beta_ \mathrm{L}$-$\beta_ \mathrm{U}$, where  $\beta_ \mathrm{L}$ is the lower bound of the threshold and  $\beta_ \mathrm{U}$ is the upper bound. Finally, the constant rate of change (0.01\% of the initial learning rate), and the direction of change (increasing or decreasing) are specified.

After determining the ideal learning rates for the S-D baseline, preliminary experiments were run where only one of the learning rates was changing, as shown in Table A.IV. In this manner, it was possible to determine what a good variable threshold is for one of the learning rates, before complicating the process by changing both learning rates at once. To further simplify the process, the rate of change was always 0.01\% of the original learning rate, and thus the different variable thresholds could be  compared easily.

Next, the best thresholds from the preliminary one-static, one-variable experiments were combined to create the DVLR experiments, as shown in Table A.V and Table A.VI. All combinations were used to determine which particular combination performed the best. Finally, as an optional step, the best thresholds from the DVLR experiments were tested with all increasing/decreasing possibilities, as shown in Table A.VII.

After all preliminary experiments were completed, the best baselines and the five experiments with the highest accuracy scores from both the one-static, one-variable tests and the DVLR tests were run over ten trials, as shown in Table A.VIII.

This process is the complete methodology used in this paper to determine the variable thresholds for the MNIST FFNN, MNIST CNN and CIFAR-10 CNN experiments. To save time, once it was discovered that the CIFAR-10 RNN had the same ideal S-S and S-D baselines as the CIFAR-10 CNN, the best DVLR thresholds from the CIFAR-10 CNN experiments were also used for CIFAR-10 RNN experiments.

\begin{figure}[t]
\centerline{\fbox{\includegraphics[scale=0.75]{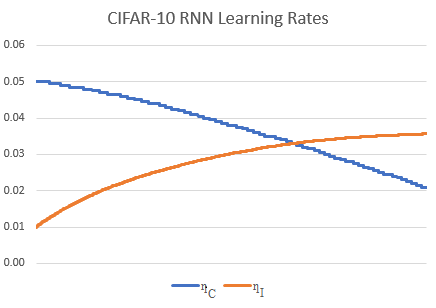}}}
\caption{Example of the learning rates obtained during one trial of the CIFAR-10 ResNet18 experiments where $\eta^0_\mathrm{C}=0.05$, VT5975-6025, 0.01\% decrease, and $\eta^0_\mathrm{I}=0.01$, VT395-405, 0.01\% increase. DVLR performed best when $\eta_\mathrm{C}$ increased, but its slope gradually decreased over time and when $\eta_ \mathrm{I}$ decreased at a nearly constant rate, demonstrating the different setup needed for the best RNN experiment as compared to the best CNN experiment. }
\end{figure}

These preliminary experiments were time consuming, but necessary. The conclusion was that MNIST FFNN and CIFAR-10 CNN perform better when $\eta_ \mathrm{C}$ is static or changes much slower than $\eta_ \mathrm{I}$. However, the MNIST CNN did not follow this general rule. To further demonstrate how the different variable thresholds affect the learning rates over time, example learning rates from one trial of the best CIFAR-10 CNN are shown in Figure 2 and example learning rates from one trial of the best CIFAR-10 RNN are shown in Figure 3. There are similarities between the two figures such as $\eta_ \mathrm{C}$ starting at a greater value than $\eta_  \mathrm{I}$ and $\eta_ \mathrm{I}$ increasing throughout the trial. However, no general rule emerges that could be used to determine ideal variable thresholds for a given network, and therefore more research is needed to discover the best practices for DVLR.

\begin{table*}[t]
\begin{center}
\caption{Accuracy of DVLR on FFNN and CNN in the MNIST domain. The results suggest that dual learning rates improve slightly over single rate, and adding variable thresholds results in further improvements. The differences are more pronounced with CNN, suggesting that DVLR scales well to more complex architectures.}
\begin{tabular}{lllll}

\textbf{Feedforward Neural Network:}                                      & \textbf{Avg Train} & \textbf{Avg Test} & \textbf{p-value} &    \\ \hline
Static-Single (S-S) Baseline, $\eta_\mathrm{S}=0.05$                              & 99.726    & 98.127   &                &    \\
Static-Dual (S-D) Baseline, $\eta_\mathrm{C}=0.075$, $\eta_\mathrm{I}=0.025$               & 99.948    & 98.245   & 0.154          &   \\ \hline
DVLR Experiments: & & & & \\
$\eta_\mathrm{C}=0.075$, $\eta^0_\mathrm{I}=0.025$ VT67-72 0.01\% inc                                  & 99.952    & 98.316   & 0.154          &   \\
$\eta^0_\mathrm{C}=0.075$ VT4995-5005 0.01\% inc, $\eta^0_\mathrm{I}=0.025$   VT195-205 0.01\% dec        & 99.952    & 98.311   & 0.154          &   \\
$\eta^0_\mathrm{C}=0.075$ VT9995-10005 0.01\% inc, $\eta^0_\mathrm{I}=0.025$ VT195-205 0.01\% inc       & 99.966    & 98.291   & 0.141          &   \\ \hline \\
\textbf{Convolutional Neural Network:} & \textbf{Avg Train} & \textbf{Avg Test} & \textbf{p-value} &    \\ \hline
Static-Single (S-S) Baseline, $\eta_\mathrm{S}=0.01$                                  & 95.000    & 98.320   &                &    \\
Static-Dual (S-D) Baseline, $\eta_\mathrm{C}=0.017$,   $\eta_\mathrm{I}=0.003$                   & 95.503    & 98.590   & 0.196          &   \\ \hline
DVLR Experiments: & & & & \\
$\eta^0_\mathrm{C}=0.017$ VT 495-505 0.01\% inc, $\eta^0_\mathrm{I}=0.003$ VT 295-305 0.01\% inc   & 95.738    & 98.777   & 0.034          & * \\
$\eta^0_\mathrm{C}=0.017$ VT 495-505 0.01\% inc, $\eta^0_\mathrm{I}=0.003$ VT 1995-2005 0.01\% dec & 95.732    & 98.744   & 0.032          & * \\
$\eta^0_\mathrm{C}=0.017$ VT 495-505 0.01\% inc, $\eta_\mathrm{I}=0.003$                         & 95.786    & 98.740   & 0.025          & * \\ \hline
\end{tabular}
\end{center}
\label{tab:mnist}
\end{table*}

\section*{Results}
The DVLR method was tested on two different databases: MNIST \cite{mnist} and CIFAR-10 \cite{cifar10}, and three different types of networks: a feedforward neural network (FFNN), a convolutional neural network (CNN \cite{convolutional}), and a residual neural network (RNN \cite{residual}). Experiments were run on networks that are best suited for each database to get the best possible baselines before experimenting with DVLR. 

\subsection*{Setup}
In the MNIST experiments, two architectures were tested. The first one was an FFNN with 500 hidden nodes, RELU activation function and batch size of 100. The second was a CNN with two convolutional layers, two fully connected linear layers, RELU activation function with dropout and a batch size of 128. In the CIFAR-10 experiment, two architectures were tested as well. The first one was a CNN with two convolutional layers, pooling layer, three fully connected linear layers, RELU activation function, and a batch size of 10. The second was a ResNet18 network that contains 16 convolutional layers, two pooling layers and utilizes a stride size of two, RELU activation function and batch size of 128. The MNIST FFNN, MNIST CNN and CIFAR-10 CNN experiments used the Adagrad optimizer while the CIFAR-10 RNN used the SGD optimizer. All networks were trained with cross-entropy loss.

The goal of the experiments was not necessarily to improve state-of-the-art, but to evaluate DVLR broadly.  These architectures were thus chosen as fundamental versions of standard neural network architectures. Also, relatively small versions of them were used to reduce computing time. It was therefore possible to run ten trials for every experiment and determine statistical significance against the S-S baseline. All code and original data can be found at https://github.com/e-liner/DVLR.

\subsection*{MNIST FFNN Results}

The average training values, testing values and \textit{t}-test \textit{p}-values of various DVLR experiments compared with the S-S baseline can be found in the first half of Table I. The differences were small, and \textit{p}-values are in the 0.15 range, so the results are suggestive only. However, many different variable thresholds were discovered that suggest improved accuracy over the baselines. The best performance was achieved when $\eta_\mathrm{C}$ was either static or slightly increasing, and when $\eta_\mathrm{I}$ was increasing or decreasing at a much higher rate.

It is important to note that the S-D baseline and the top three DVLR experiments performed better than the S-S baseline. This result demonstrates that the dual learning rate method can potentially increase the accuracy of a simple feedforward network. Additionally, all three DVLR experiments performed better than both the S-S and S-D baselines, which demonstrates the potential value of the variable threshold update method.

\subsection*{MNIST CNN Results}

The average training values, testing values and \textit{t}-test \textit{p}-values of various DVLR experiments compared with the S-S baseline can be found in the second half of Table I.  The $p$- value for the S-D baseline is 0.196, however all DVLR experiments were found to be statistically significantly better than the baseline.
One particularly powerful variable threshold was discovered for $\eta_\mathrm{C}$ that improved significantly over the baseline. The thresholds for $\eta_\mathrm{I}$ vary widely, and utilize variable thresholds that are static, slightly changing, and changing at a much higher rate.

These results suggest that the dual learning rate method provides an advantage compared to a standard single learning rate, but that both the dual learning rate method and variable threshold update method had to be utilized to obtain a significant increase. The improvements of the MNIST CNN results are more pronounced than the MNIST FFNN results, suggesting that DVLR should scale up well to more complex architectures.

\begin{table*}[t]
\begin{center}
\caption{Accuracy of DVLR on CNN and RNN in the CIFAR-10 domain. The results again suggest that dual learning rates improve slightly over single rate, and adding variable thresholds results in further improvements. The differences are less pronounced on RNN, presumably because the ratios from CNN were used for it instead of customizing them to the architecture. However, although the differences are small, they are remarkably consistent across all comparisons, suggesting that DVLR is a promising technique for improving neural network training.}
\begin{tabular}{lllll}
\textbf{Convolutional Neural Network:} & \textbf{Avg Train} & \textbf{Avg Test} & \textbf{p-value} &    \\ \hline
Static-Single (S-S) Baseline, $\eta_\mathrm{S}=0.03$                            & 66.326    & 61.139   &         &   \\
Static-Dual (S-D) Baseline, $\eta_\mathrm{C}=0.05$,   $\eta_\mathrm{I}=0.01$                 & 68.062    & 62.434   & 0.361   &   \\ \hline
DVLR Experiments: & & & & \\
$\eta_\mathrm{C}=0.05$, $\eta^0_\mathrm{I}=0.01$ VT175-225, 0.01\% inc                         & 70.292    & 63.285   & 0.126   &  \\
$\eta^0_\mathrm{C}=0.05$ VT6975-7025 0.01\% inc, $\eta^0_\mathrm{I}=0.01$   VT395-405 0.01\% inc & 69.706    & 63.214   & 0.171   &  \\
$\eta^0_\mathrm{C}=0.05$ VT5975-6025 0.01\% dec, $\eta^0_\mathrm{I}=0.01$   VT395-405 0.01\% inc & 69.552    & 63.158   & 0.163   &  \\ \hline \\
\textbf{Residual Neural Network:} & \textbf{Avg Train} & \textbf{Avg Test} & \textbf{p-value} &    \\ \hline
Static-Single (S-S) Baseline, $\eta_\mathrm{S}=0.03$                            & 91.416    & 72.727   &         &   \\
Static-Dual (S-D) Baseline, $\eta_\mathrm{C}=0.05$,   $\eta_\mathrm{I}=0.01$                 & 92.066    & 74.353   & 0.391   &   \\ \hline
DVLR Experiments: & & & & \\
$\eta^0_\mathrm{C}=0.05$ VT5975-6025 0.01\% dec, $\eta^0_\mathrm{I}=0.01$   VT395-405 0.01\% inc & 96.085    & 74.957   & 0.334   &   \\
$\eta^0_\mathrm{C}=0.05$ VT6975-7025 0.01\% inc, $\eta^0_\mathrm{I}=0.01$   VT395-405 0.01\% inc & 90.319    & 74.637   & 0.402   &   \\
$\eta_\mathrm{C}=0.05$, $\eta^0_\mathrm{I}=0.01$ VT195-205, 0.01\% inc                         & 92.173    & 74.387   & 0.371   &   \\ \hline
\end{tabular}
\end{center}
\label{tab:cifar}
\end{table*}
\subsection*{CIFAR-10 CNN Results}

The average training values, testing values, and \textit{t}-test \textit{p}-values of various DVLR experiments compared with the S-S baseline can be found in the first half of Table II. The S-D baseline was not found to be significantly different from S-S, however the top three DVLR experiments were close to significant in the $p=0.12$ to 0.17 range. Many different variable thresholds were found that resulted in potentially improved accuracy from the baselines. The $\eta_\mathrm{C}$ performed best when it was either static or slightly changing, and $\eta_\mathrm{I}$ performed best when it was increasing at a much higher rate, similar to the MNIST FFNN results.

In CIFAR-10, the CNN accuracy increases more than in MNIST, suggesting that the method scales well to larger datasets. DVLR also improves upon the S-D baseline, suggesting that the dual learning rate method is stronger when used with the variable threshold update method.

\subsection*{CIFAR-10 RNN Results}
The average training values, testing values, and \textit{t}-test \textit{p}-values of various DVLR versions compared with the S-S baseline can be found in the second half of Table II. With this architecture, neither the S-D baseline, nor the DVLR experiments were found to be significantly different from S-S. This result may be due to the lack of empirical testing to determine the best variable thresholds for this particular network. More specifically, in the interest of time, once it was discovered that the CIFAR-10 RNN had the same ideal S-S and S-D baselines as the CIFAR-10 CNN, the best variable ratios from the CIFAR-10 CNN experiments were also used on the CIFAR-10 RNN. Using the the full methodology to obtain new variable ratios may be necessary to obtain significantly better results.

\section*{Discussion and Future Work}

Statistical difference is rarely estimated with modern deep learning architectures due to the excessive computational cost: Training of a full-scale model on a large dataset can have a carbon footprint of several cars \cite{carbon}. The approach taken in this paper was to scale down the models and the datasets to the level where such repetitions could be done, in order to evaluate the DVLR technique more comprehensively. Although the improvements are small and not always statistically significant, they are remarkably consistent: across all four comparisons, S-D is always better than S-S, and across all twelve comparisons, DVLR is always better than S-D. The experiments thus provide substantial evidence that the DVLR technique is effective. Because the comparisons span multiple architectures and multiple domains, they suggest that DVLR can be used widely with neural networks that use gradient descent as their update method.  Also, as the networks and domains increase in size and complexity, DVLR is likely to have a more pronounced effect. It may therefore constitute a robust and general technique for the modern machine learning toolbox.

A significant aspect of DVLR is its Behavioral Psychology motivation, and there are intriguing connections to neuroscience as well. In the brain, amygdala and ventral striatum work together to facilitate reinforcement learning \cite{averbeck}. Amygdala has a faster learning rate than the ventral striatum; having multiple neural systems learn at different rates may thus facilitate more effective learning in dynamic environments. It may be possible to analyze the biological data on learning rates in more depth, and refine the variable learning rate methods of DVLR further.

\section*{Conclusion}
DVLR is a new training technique for neural networks that is motivated by behavioral psychology. DVLR is a combination of two contributions. First, it takes advantage of dual learning rates, $\eta_\mathrm{C}$ and $\eta_\mathrm{I}$, that correspond to the network's correct and incorrect responses. Second, it demonstrates that their impact is increased with variable threshold update schedules. DVLR was tested on feedforward networks and convolutional networks with the MNIST dataset and on convolutional networks and residual networks with the CIFAR-10 dataset. The experiments suggest a consistent improved accuracy on all three network types over both domains. Moreover, it was found to be more powerful in larger architectures and datasets, making it a promising technique for the general machine learning toolbox.

\appendix
The tables below give details on preliminary experiments for configuring the learning rates and threshold ranges for the CIFAR-10 CNN experiments. The MNIST FFNN and MNIST CNN experiments were also determined in this manner.

\begin{table}[h]
\caption{CIFAR-10 CNN Baseline}
\vspace*{-4ex}
\begin{center}
\begin{tabular}{lll}
Static-Single (S-S)   Baselines:                             & Test Avg.    \\ \hline
$\eta_\mathrm{S}=0.0001$                                                        & 27.800   \\
$\eta_\mathrm{S}=0.0005$                                                        & 36.873     \\
$\eta_\mathrm{S}=0.001$                                                      & 42.820     \\
$\eta_\mathrm{S}=0.005$                                                   & 52.967     \\
$\eta_\mathrm{S}=0.01$                                                          & 59.493     \\
$\eta_\mathrm{S}=0.02$                                                          & 60.617     \\
$\eta_\mathrm{S}=0.03$                                                          & 62.207   \\
$\eta_\mathrm{S}=0.04$                                                          & 61.907 \\
$\eta_\mathrm{S}=0.05$                                                          & 59.947     \\
$\eta_\mathrm{S}=0.06$                                                     & 58.950    \\
\\
Static-Dual (S-D)   Baselines:                               & Test Avg.   \\ \hline
$\eta_\mathrm{C}=0.035$, $\eta_\mathrm{I}=0.025$                               & 61.897    \\
$\eta_\mathrm{C}=0.04$, $\eta_\mathrm{I}=0.02$                                               & 61.270    \\
$\eta_\mathrm{C}=0.05$, $\eta_\mathrm{I}=0.01$                                               & 63.560    \\
$\eta_\mathrm{C}=0.025$, $\eta_\mathrm{I}=0.035$                                             & 61.030    \\
$\eta_\mathrm{C}=0.02$, $\eta_\mathrm{I}=0.04$                                               & 60.527    \\
$\eta_\mathrm{C}=0.01$, $\eta_\mathrm{I}=0.05$                                               & 56.047    \\
$\eta_\mathrm{C}=0.05$, $\eta_\mathrm{I}=0.03$                                               & 60.667    \\
$\eta_\mathrm{C}=0.03$, $\eta_\mathrm{I}=0.01$                                               & 62.330    \\
$\eta_\mathrm{C}=0.05$, $\eta_\mathrm{I}=0.02$                                               & 61.010    \\
$\eta_\mathrm{C}=0.04$, $\eta_\mathrm{I}=0.03$                                               & 61.630    \\
\end{tabular}
\end{center}
\label{tab:my-table1}
\end{table}

\begin{table}[h]
\caption{CIFAR-10 CNN One Static, One Variable Tests}
\vspace*{-3ex}
\begin{center}
\begin{tabular}{lll}
Method:                      &  Test Avg.         \\ \hline
$\eta_\mathrm{C}=0.05$, $\eta^0_\mathrm{I}=0.01$ VT95-105, 0.01\%   inc                 & 60.243    \\
$\eta_\mathrm{C}=0.05$, $\eta^0_\mathrm{I}=0.01$ VT195-205,   0.01\% inc                & 63.377    \\
$\eta_\mathrm{C}=0.05$, $\eta^0_\mathrm{I}=0.01$ VT150-250,   0.01\% inc                & 62.567    \\
$\eta_\mathrm{C}=0.05$, $\eta^0_\mathrm{I}=0.01$ VT175-225,   0.01\% inc                & 64.120  \\
$\eta_\mathrm{C}=0.05$, $\eta^0_\mathrm{I}=0.01$ VT295-305,   0.01\% inc                & 62.287    \\
$\eta_\mathrm{C}=0.05$, $\eta^0_\mathrm{I}=0.01$ VT495-505,   0.01\% inc                & 62.837    \\
$\eta_\mathrm{C}=0.05$, $\eta^0_\mathrm{I}=0.01$ VT995-1005,   0.01\% inc               & 63.097    \\
$\eta_\mathrm{C}=0.05$, $\eta^0_\mathrm{I}=0.01$   VT1150-1350, 0.01\% inc              & 62.503    \\
$\eta_\mathrm{C}=0.05$, $\eta^0_\mathrm{I}=0.01$ VT95-105, 0.01\%   dec                 & 61.853    \\
$\eta_\mathrm{C}=0.05$, $\eta^0_\mathrm{I}=0.01$ VT195-205,   0.01\% dec                & 61.023    \\
$\eta_\mathrm{C}=0.05$, $\eta^0_\mathrm{I}=0.01$ VT295-305,   0.01\% dec                & 63.247    \\
$\eta_\mathrm{C}=0.05$, $\eta^0_\mathrm{I}=0.01$ VT495-505,   0.01\% dec                & 62.257    \\
$\eta_\mathrm{C}=0.05$, $\eta^0_\mathrm{I}=0.01$ VT2475-2525,   0.01\% dec              & 63.170   \\ \hline
$\eta^0_\mathrm{C}=0.05$ VT95-105 0.01\% dec, $\eta_\mathrm{I}=0.01$                           & 53.063     \\
$\eta^0_\mathrm{C}=0.05$ VT195-205 0.01\% dec, $\eta_\mathrm{I}=0.01$                          & 53.343     \\
$\eta^0_\mathrm{C}=0.05$ VT295-305 0.01\% dec, $\eta_\mathrm{I}=0.01$                          & 54.357     \\
$\eta^0_\mathrm{C}=0.05$ VT495-505 0.01\% dec, $\eta_\mathrm{I}=0.01$                          & 54.480     \\
$\eta^0_\mathrm{C}=0.05$ VT1475-2525 0.01\%   dec, $\eta_\mathrm{I}=0.01$                      & 60.777     \\
$\eta^0_\mathrm{C}=0.05$ VT3975-4075 0.01\% dec, $\eta_\mathrm{I}=0.01$                        & 63.270    \\
$\eta^0_\mathrm{C}=0.05$ VT5975-6025 0.01\% dec, $\eta_\mathrm{I}=0.01$                        & 63.157    \\
$\eta^0_\mathrm{C}=0.05$ VT6975-7025 0.01\% dec, $\eta_\mathrm{I}=0.01$                        & 62.907    \\
$\eta^0_\mathrm{C}=0.05$ VT95-105 0.01\% inc, $\eta_\mathrm{I}=0.01$                          & 49.017     \\
$\eta^0_\mathrm{C}=0.05$ VT195-205 0.01\% inc, $\eta_\mathrm{I}=0.01$                          & 55.717    \\
$\eta^0_\mathrm{C}=0.05$ VT295-305 0.01\% inc, $\eta_\mathrm{I}=0.01$                          & 58.117     \\
$\eta^0_\mathrm{C}=0.05$ VT495-505 0.01\% inc, $\eta_\mathrm{I}=0.01$                          & 60.107    \\
\end{tabular}
\end{center}
\label{tab:my-table2}
\end{table}

\begin{table}[h]
\caption{CIFAR-10 CNN Best from One Static, One Variable Tests}
\vspace*{-3ex}
\begin{center}
\begin{tabular}{lll}
Best $\eta_\mathrm{I}$ changes:                                                 &    Test Avg.   \\ \hline
$\eta_\mathrm{C}=0.05$, $\eta^0_\mathrm{I}=0.01$ VT175-225,   0.01\% inc                & 64.120   \\
$\eta_\mathrm{C}=0.05$, $\eta^0_\mathrm{I}=0.01$ VT195-205,   0.01\% inc                & 63.377   \\
$\eta_\mathrm{C}=0.05$, $\eta^0_\mathrm{I}=0.01$ VT295-305,   0.01\% dec                & 63.247   \\
$\eta_\mathrm{C}=0.05$, $\eta^0_\mathrm{I}=0.01$ VT2475-2525,   0.01\% dec              & 63.170   \\
$\eta_\mathrm{C}=0.05$, $\eta^0_\mathrm{I}=0.01$ VT995-1005,   0.01\% inc               & 63.097   \\
$\eta_\mathrm{C}=0.05$, $\eta^0_\mathrm{I}=0.01$ VT495-505,   0.01\% inc                & 62.837   \\
$\eta_\mathrm{C}=0.05$, $\eta^0_\mathrm{I}=0.01$ VT150-250,   0.01\% inc                & 62.567   \\
$\eta_\mathrm{C}=0.05$, $\eta^0_\mathrm{I}=0.01$ VT1150-1350,   0.01\% inc              & 62.503   \\
                                                                 &           \\
Best $\eta_\mathrm{C}$ changes:                                                 &   Test Avg.         \\ \hline
$\eta^0_\mathrm{C}=0.05$ VT3975-4075 0.01\% dec, $\eta_\mathrm{I}=0.01$               & 63.270      \\
$\eta^0_\mathrm{C}=0.05$ VT5975-6025 0.01\% dec, $\eta_\mathrm{I}=0.01$               & 63.157      \\
$\eta^0_\mathrm{C}=0.05$ VT6975-7025 0.01\% dec, $\eta_\mathrm{I}=0.01$               & 62.907      \\
\end{tabular}
\end{center}
\label{tab:my-table3}
\end{table}

\begin{table}[h]
\caption{CIFAR-10 CNN DVLR Tests}
\vspace*{-5ex}
\begin{center}
\begin{tabular}{lll}
Method:                                      &   Test Avg.       \\ \hline
$\eta^0_\mathrm{C}=0.05$ VT3975-4075 dec, $\eta^0_\mathrm{I}=0.01$   VT175-225 inc & 60.597      \\
$\eta^0_\mathrm{C}=0.05$ VT3975-4075 dec, $\eta^0_\mathrm{I}=0.01$   VT195-205 inc & 61.703      \\
$\eta^0_\mathrm{C}=0.05$ VT3975-4075 dec, $\eta^0_\mathrm{I}=0.01$   VT295-305 dec & 62.650      \\
$\eta^0_\mathrm{C}=0.05$ VT3975-4075 dec, $\eta^0_\mathrm{I}=0.01$   VT395-405 inc & 62.653      \\ \hline
$\eta^0_\mathrm{C}=0.05$ VT5975-6025 dec, $\eta^0_\mathrm{I}=0.01$   VT175-225 inc & 62.790      \\
$\eta^0_\mathrm{C}=0.05$ VT5975-6025 dec, $\eta^0_\mathrm{I}=0.01$   VT195-205 inc & 63.157      \\
$\eta^0_\mathrm{C}=0.05$ VT5975-6025 dec, $\eta^0_\mathrm{I}=0.01$   VT295-305 dec & 61.850      \\
$\eta^0_\mathrm{C}=0.05$ VT5975-6025 dec, $\eta^0_\mathrm{I}=0.01$   VT395-405 inc & 63.943      \\ \hline
$\eta^0_\mathrm{C}=0.05$ VT6975-7025 dec, $\eta^0_\mathrm{I}=0.01$   VT175-225 inc & 62.087      \\
$\eta^0_\mathrm{C}=0.05$ VT6975-7025 dec, $\eta^0_\mathrm{I}=0.01$   VT195-205 inc & 63.237      \\
$\eta^0_\mathrm{C}=0.05$ VT6975-7025 dec, $\eta^0_\mathrm{I}=0.01$   VT295-305 dec & 61.283      \\
$\eta^0_\mathrm{C}=0.05$ VT6975-7025 dec, $\eta^0_\mathrm{I}=0.01$   VT395-405 inc & 63.187      \\
\end{tabular}
\end{center}
\label{tab:my-table4}
\end{table}

\begin{table*}[t]
\begin{minipage}{\textwidth}
\hspace*{-4ex}
\begin{minipage}{0.5\columnwidth}
\caption{CIFAR-10 CNN DVLR Tests, all directions}
\vspace*{-5ex}
\begin{center}
\begin{tabular}{lll}
Method:                         &  Test Avg.          \\ \hline
$\eta^0_\mathrm{C}=0.05$ VT5975-6025 dec, $\eta^0_\mathrm{I}=0.01$ VT395-405 inc & 63.158      \\
$\eta^0_\mathrm{C}=0.05$ VT5975-6025 dec, $\eta^0_\mathrm{I}=0.01$ VT395-405 dec & 61.724   \\
$\eta^0_\mathrm{C}=0.05$ VT5975-6025 inc, $\eta^0_\mathrm{I}=0.01$ VT395-405 inc & 62.324    \\
$\eta^0_\mathrm{C}=0.05$ VT5975-6025 inc, $\eta^0_\mathrm{I}=0.01$ VT395-405 dec & 62.517    \\
$\eta^0_\mathrm{C}=0.05$ VT6975-7025 dec, $\eta^0_\mathrm{I}=0.01$ VT395-405 inc & 63.092    \\
$\eta^0_\mathrm{C}=0.05$ VT6975-7025 dec, $\eta^0_\mathrm{I}=0.01$ VT395-405 dec & 62.402    \\
$\eta^0_\mathrm{C}=0.05$ VT6975-7025 inc, $\eta^0_\mathrm{I}=0.01$ VT395-405 inc & 63.214      \\
$\eta^0_\mathrm{C}=0.05$ VT6975-7025 inc, $\eta^0_\mathrm{I}=0.01$ VT395-405 dec & 62.053    \\\end{tabular}
\end{center}
\label{tab:my-table5}
\end{minipage}
\hfill
\begin{minipage}{0.5\columnwidth}
\hspace*{-2ex}
\vspace*{-3ex}
\begin{center}
\caption{CIFAR-10 CNN Baselines and Experiments, 10-set trials}
\vspace*{-1ex}
\begin{tabular}{ll}
Method & Test Avg. \\ \hline
Static-Singular {S-S} Baseline $\eta_\mathrm{S}=0.03$                          & 61.139 \\
Static-Dual (S-D) Baseline $\eta_\mathrm{C}=0.05$, $\eta_\mathrm{I}=0.01$                   & 62.434 \\
$\eta_\mathrm{C}=0.05$, $\eta^0_\mathrm{I}=0.01$ VT175-225 inc                & 63.285 \\
$\eta^0_\mathrm{C}=0.05$ VT6975-7025 inc, $\eta^0_\mathrm{I}=0.01$  VT395-405 inc & 63.214 \\
$\eta^0_\mathrm{C}=0.05$ VT5975-6025 dec, $\eta^0_\mathrm{I}=0.01$  VT395-405 inc & 63.158 \\
$\eta_\mathrm{C}=0.05$, $\eta^0_\mathrm{I}=0.01$ VT195-205, inc                & 63.151 \\
$\eta^0_\mathrm{C}=0.05$ VT6975-7025 dec, $\eta^0_\mathrm{I}=0.01$  VT395-405 inc & 63.092
\end{tabular}
\end{center}
\label{tab:my-table6}
\end{minipage}
\end{minipage}
\vspace*{150ex}
\end{table*}

\end{document}